\documentclass[10pt,twocolumn,letterpaper]{article}

\usepackage{iccv}
\usepackage{times}
\usepackage{epsfig}
\usepackage{graphicx}
\usepackage{amsmath}
\usepackage{amssymb}

\usepackage{booktabs}
\usepackage{bm}
\usepackage{times}
\usepackage{epsfig}
\usepackage{float} 
\usepackage{makecell}
\usepackage{colortbl}
\usepackage{mathtools}
\usepackage{extarrows}
\usepackage{algorithmic}
\usepackage{adjustbox}
\usepackage[percent]{overpic}
\usepackage[linesnumbered,ruled]{algorithm2e}
\SetKwComment{Comment}{$\triangleright$\ }{}
\usepackage{balance}
\usepackage{stfloats}
\usepackage{textcomp}

\usepackage{multirow}
\usepackage{comment}
\newcommand*\samethanks[1][\value{footnote}]{\footnotemark[#1]}

\graphicspath{{SR_pruning/}}

\usepackage[pagebackref=true,breaklinks=true,letterpaper=true,colorlinks,bookmarks=false]{hyperref}
\usepackage[dvipsnames]{xcolor}
\hypersetup{
    colorlinks=true,       
    linkcolor=red,          
    citecolor=Green,        
}
\usepackage{colortbl}
\definecolor{cYellow}{HTML}{FFFFCC}
\definecolor{cRed}{HTML}{FFCCCC} 
\definecolor{cGrey}{HTML}{E5E4E2}

\iccvfinalcopy 

\def\iccvPaperID{3630} 

\ificcvfinal\fi

\begin{document}

\title{Iterative Soft Shrinkage Learning for Efficient Image Super-Resolution} 
%

\author{
Jiamian Wang$^{1}$,
Huan Wang$^{2}$,
Yulun Zhang$^{3}$\thanks{Corresponding authors: Yulun Zhang (yulun100@gmail.com) and Zhiqiang Tao (zxtics@rit.edu)},
Yun Fu$^{2}$,
and Zhiqiang Tao$^{1}$\samethanks
\\
$^{1}$Rochester Institute of Technology, $^{2}$Northeastern University, 
$^{3}$ETH Z\"{u}rich\\
}

\maketitle
\ificcvfinal\thispagestyle{empty}\fi

   
\begin{abstract}
   Image super-resolution (SR) has witnessed extensive neural network designs from CNN to transformer architectures. 
   However, prevailing SR models suffer from prohibitive memory footprint and intensive computations, which limits further deployment on edge devices. 
   This work investigates the potential of network pruning for super-resolution to take advantage of off-the-shelf network designs and reduce the underlying computational overhead.
   Two main challenges remain in applying pruning methods for SR.
   First, the widely-used filter pruning technique reflects limited granularity and restricted adaptability to diverse network structures.
   Second, existing pruning methods generally operate upon a pre-trained network for the sparse structure determination, hard to get rid of dense model training in the traditional SR paradigm. To address these challenges, we adopt unstructured pruning with sparse models directly trained from scratch. Specifically, we propose a novel Iterative Soft Shrinkage-Percentage (ISS-P) method by optimizing the sparse structure of a randomly initialized network at each iteration and tweaking unimportant weights with a small amount proportional to the magnitude scale on-the-fly.  
   We observe that the proposed ISS-P can dynamically learn sparse structures adapting to the optimization process and preserve the sparse model's trainability by yielding a more regularized gradient throughput. Experiments on benchmark datasets demonstrate the effectiveness of the proposed ISS-P  over diverse network architectures.
   Code is available at \url{https://github.com/Jiamian-Wang/Iterative-Soft-Shrinkage-SR}

\end{abstract}

\section{Introduction}\label{sec: introduction}
Single image super-resolution~\cite{johnson2016perceptual,kim2016deeply} aims to reconstruct the high-resolution (HR) image from a low-resolution (LR) input. 
Towards a high-fidelity reconstruction,  research efforts have been made by relying on the strong modeling capacity of convolutional neural networks~\cite{dong2014learning,kim2016accurate,lim2017enhanced,zhang2018image}. More recently, advanced Transformer architectures~\cite{liang2021swinir,zamir2022restormer,zheng2022cross} are elaborated, enabling photorealistic restoration. 
 Despite the impressive performance, the excessive memory footprint of existing models has been \emph{de facto} in the field, which inevitably prohibits the deployment of advanced SR models on computational-constrained devices. 

To alleviate the computational complexity, we study network pruning, which takes advantage of off-the-shelf advanced network architectures to realize efficient yet accurate SR models. Network pruning has been long developed in two mainstream directions. On the one hand, filter pruning (structured pruning)~\cite{li2016pruning} cuts off the specific filter of convolutional layers, among which representative practices in SR is to prune cross-layer filters governed by residual connections~\cite{zhang2021aligned,zhang2021learning}. However, these methods need to consider the layer-wise topology by posing structural constraints, requiring a heuristic design and thus making them inflexible. 
Plus, filter pruning inhibits a more fine-grained manipulation to the network.
On the other hand, weight pruning (unstructured pruning) directly removes weight scalars across the network, endowed with more flexibility by accommodating weight discrepancy. Also, the weight pruning method allows a very high pruning ratio, \emph{e.g.}, a ratio of $99\%$ with competitive performance~\cite{han2015deep,han2015learning}. To this end, this work focuses on delivering a highly-adaptive unstructured pruning solution for diverse SR architectures.

Generally, pruning algorithms are widely recognized to have three steps: (1) pre-training, (2) sparse structure acquisition, and (3) fine-tuning the sparse network. Among these steps, the dense network pre-training usually introduces heavy costs beyond the sparse network optimization. For example, before obtaining a sparse network, the CAT~\cite{zheng2022cross} network architecture takes $2$ days to train its dense counterpart on $4$ A100 GPUs. Thus, a natural question arises to save training time further -- can we directly explore the sparsity of network structures from random initialization?

We start from the baseline method by performing random pruning on weights at initialization, whose limitation  
is the irrelevance between the sparse structure and the weight distribution varying to the optimization. Following this line, we apply the widely-used $L_{1}$ norm~\cite{li2016pruning,han2015learning} pruning on randomly initialized weights. 
However, the immutable sparsity cannot be well aligned with the optimization, 
leading to limited performance. To tackle this problem, we introduce an iterative hard thresholding~\cite{blumensath2009iterative,candes2006robust} method (IHT) stemming from compressive sensing~\cite{candes2006stable,donoho2006compressed}, where the iterative gradient descent step is regularized by a hard thresholding function. Unlike previous works, we tailor IHT to iteratively set unimportant weights as zeros and preserve the important weight magnitudes. 
By this means, the sparse structure adapts to the weight distribution throughout the training, which potentially better selects essential weights. 
However, the sparse structure alignment in IHT is heavily susceptible to the magnitude-gradient relationship. 
The zeroed weight can be continually trapped as ``unimportant'' once the scales of magnitude and gradient are incomparable. Moreover, by directly zeroing out unimportant weights, IHT blocks the error back-propagation at each iteration, especially hindering the optimization in shallow layers.

To address the aforementioned negative effects, we introduce a more flexible thresholding function for an expressive treatment of unimportant weights. A natural tuning approach is to softly shrink the weights rather than hard threshold. 
We first explore the soft shrinkage by a growing regularization method~\cite{zhang2021learning}, namely ISS-R. Per each iteration, the proposed ISS-R constrains weight magnitudes with a gradually increasing weighted $L_2$ regularization, to avoid the conflict between network pruning and smooth sparse network optimization.  
However, the growing regularization schedule involves a number of hyperparameter tuning, requiring cumbersome manual efforts.  
Notably, the $L_2$ regularization shrinkage inside ISS-R is, in essence, proportional to the weight magnitude. Based on this insight, we propose a new iterative soft shrinkage function to simplify the regularization by equivalently shrinking the weight with a percentage (ISS-P).  It turns out that ISS-P not only encourages dynamic network sparsity, but also preserves the sparse network trainability, resulting in better convergence.
We summarize the contributions of this work as follows:
\begin{itemize}
    \setlength\itemsep{0em}
    \item 
    We introduce a novel unstructured pruning method, namely iterative soft shrinkage-percentage (ISS-P), which is compatible with diverse SR network designs. Unlike existing pruning strategies for SR, the proposed method trains the sparse network from scratch, providing a practical solution for sparse network acquisition under computational budget constraints. 
    \vspace{-1mm}
    
    \item 
    We explore pruning behaviors by interpreting the trainability of sparse networks. The proposed ISS-P enjoys a more promising gradient convergence and enables dynamic sparse structures in the training process, offering new insights to design pruning methods for SR. 
    
    \item 
    Extensive experimental results on benchmark testing datasets at different pruning ratios and scales demonstrate the effectiveness of the proposed method compared with state-of-the-art pruning solutions. 
    
\end{itemize}

\section{Related Work}\label{sec: related work}

\noindent \textbf{Single Image Super-Resolution}.
The task of single image super-resolution has been developed with remarkable progress since the first convolutional network of  SRCNN~\cite{dong2014learning} was introduced. By taking advantage of the residual structure~\cite{He2016DeepRL}, VDSR~\cite{kim2016accurate} further encourages fine-grained reconstruction at rich textured areas. Based on it, EDSR~\cite{lim2017enhanced} witnessed a promotion by empowering the regression with deeper network depth and a simplified structure.  Besides, RCAN~\cite{zhang2018image} outperforms its counterparts by incorporating channel attention into the residual structure. 
HPUN~\cite{sun2023hybrid} proposes a downsampling module for a efficient modeling.
More recently, transformer~\cite{dosovitskiy2020image,vaswani2017attention} has become a prevailing option due to its long-range dependency modeling capacity. SwinIR~\cite{liang2021swinir} equips attention with spatial locality and translation invariance properties. Another design of CAT~\cite{zheng2022cross} exploiting the power of the transformer by developing a flexible window interaction. However, advanced SR models are characterized by rising computational overhead and growing storage costs.

\noindent \textbf{Neural Network Pruning in SR}.
Neural network pruning~\cite{reed1993pruning,sze2017efficient} compresses and accelerates the network by removing redundant parameters. It has been developed in two categories: (1) Structured pruning, which mainly refers to the filter pruning~\cite{he2020learning,he2018soft,he2017channel,lebedev2016fast,li2016pruning,li2020group,lin2020hrank,wen2016learning}, removes the redundant filters for a sparsity pattern exploitation. Recently, two novel works discussed the filter pruning specialized for SR models. ASSL~\cite{zhang2021aligned} handles the residual network by regularizing the pruned filter locations of different layers upon an alignment penalty. GASSL~\cite{10106130} expands the ASSL by a Hessian-Aided Regularization. Later, SRP~\cite{zhang2021learning} makes a step further by simplifying the determination of pruned filter indices and yields a state-of-the-art performance. However, both of them require heuristic design for the pruning schedule, hard to extend to diverse neural architectures. (2) Unstructured pruning (weight pruning)~\cite{han2015learning} directly manipulates the weights for the sparsity determination. Despite the flexibility, there lacks an effective pruning strategy proposed to broadly handle advanced SR networks. Our work is to deliver a more generalized solution for different architectures.  
Besides our setting, another emerging trend is to develop fine-grained pruning upon N:M sparsity~\cite{hubara2021accelerated,mishra2021accelerating}. Among them, SLS~\cite{oh2022attentive} adapts the layer-wise sparsity level upon the trade-off between the computational cost and performance for the convolutional network pruning. Yet, the effectiveness of this method toward novel neural architectures, \emph{e.g.}, Transformers, has not been explored.

\section{Method}\label{sec: method}
We give the background of SR in Section~\ref{subsec: SR} and pruning prerequisites in Section~\ref{subsec: prerequisites}. We then tailor the classic pruning method to SR by iterative hard thresholding (IHT) in Section~\ref{subsec: IHT}. We develop our method of iterative soft shrinkage by percentage (ISS-P) in Section~\ref{subsec: ISS}.

\subsection{Single Image Super-resolution}
\label{subsec: SR}
The task of single image super-resolution is to restore the high-resolution (HR) image $I_{\texttt{HR}}$ upon the low-resolution (LR) counterpart $I_{\texttt{LR}}$ as $I_{\texttt{HR}} = F(\Theta; I_{\texttt{LR}})$, 
where $F(\cdot)$ is the SR network and $\Theta$ denotes all of the learnable parameters in the network. Given a training dataset $\mathcal{D}$, privileging practice is to formulate the SR as a pixel-wise reconstruction problem and solve it with the MSE loss by 
\begin{equation}\label{eq: SR loss}
\begin{aligned}
    J(\Theta;\mathcal{D})= \frac{1}{|\mathcal{D}|}\sum_{\mathcal{D}}||F(\Theta;I_{\texttt{LR}})-I_{\texttt{HR}})||^2, 
\end{aligned}
\end{equation}
where $I_{\texttt{LR}},I_{\texttt{HR}}\in\mathcal{D}$.
While existing deep SR networks have achieved an impressive photorealistic performance, their cumbersome computation inhibits further deployment on edge devices. In this work, we propose a generalized pruning method for off-the-shelf network designs. We introduce the prerequisites for pruning in the following. 

\subsection{Prerequisites for Pruning}
\label{subsec: prerequisites}
\noindent \textbf{Pruning Granularity}. Pruning granularity refers to the basic weight group to be removed from the network. In unstructured (weight) pruning, weight scalars are taken as manipulation units, which allows different treatment toward the neighbored weights. Besides, recent SR models have been advanced with diverse operators and structures, \emph{e.g.}, convolution, multi-head self-attention, residual blocks, etc. Unstructured pruning can be flexibly incorporated into diverse structures without additional constraints. 

\noindent \textbf{Pruning Schedule}. The prevailing pruning schedule is widely recognized as three steps: (1) pre-training a dense network, (2) pruning, and (3) fine-tuning the sparse network for performance compensation. However, training a dense SR network in step (1) from scratch already introduces heavy costs beyond the sparse network optimization in step (2$\sim$3).
To alleviate this problem, in this work, we exploit the sparse network acquisition schedule directly from networks at random initialization and get rid of the first step. 

\noindent \textbf{Baseline Methods}. Bearing with above considerations, we firstly apply several baseline pruning methods to randomly initialized SR networks, including 1) directly training a sparse network with a random sparse structure, namely scratch, and 2) $L_{1}$ norm pruning (dubbed as $L_{1}$-norm)~\cite{li2016pruning}. These two baselines, Sctrach and $L_{1}$-norm, are widely used in mainstream pruning literature. However,  they both fail to adjust the sparse structure adapting to the weight magnitude fluctuation incurred by gradient descent -- initially unimportant weights with small magnitude can be finally preserved due to negligible gradient at certain iterations of backpropagation, while initially important weights (large magnitude) with prominent gradients can be eliminated. 

\noindent\textbf{Notations}. Let $k\in K$ be the training iterations, where $K$ is consistent among different pruning methods. Pruning is conducted in a layer-wise manner in each iteration. Given a network with $L$ layers, we define $\theta^{[k]}_l \in \Theta$ as an arbitrary weight magnitude in the $l$-th layer at $k$-th iteration. Without losing the generality, we will present pruning by taking the $\theta^{[k]}_l$ as an example throughout the methodology.  

\subsection{Iterative Hard Thresholding}
\label{subsec: IHT}
To better capture essential network weights during the optimization, we introduce an iterative hard thresholding (IHT) method in light of compressive sensing (CS)~\cite{candes2006stable,donoho2006compressed}. Typically, IHT operates on the iterative gradient descent with a \emph{hard thresholding} function, serving as a widely-used method for $L_0$-norm-based non-convex optimization problems~\cite{candes2006robust}.
Unlike the IHT practices in CS, we develop a hard thresholding function $H(\cdot)$ to adjust the weight magnitudes, which takes effect at each forward propagation by 
\begin{small}
\begin{equation}
    \begin{aligned}
    \theta^{[k]}_l \!=\! H(\theta^{[k]}_l) ~~ \textup{where}~~
        H(\theta^{[k]}_l)\!=\!\begin{cases} \theta^{[k]}_l,  \textup{~if~} \theta^{[k]}_l\geq\tau^{[k]}_l,\\ 0, \textup{~if~} \theta^{[k]}_l<\tau^{[k]}_l, \end{cases}
    \end{aligned}
\end{equation}
\end{small}where $\tau^{[k]}_l$ denotes the threshold magnitude of the $l$-th layer at $k$-th iteration, determined by an $L_1$-norm sorting of the $l$-th layer weights with a given pruning ratio $r$.  
We define pruning iterations as $K_{\texttt{p}}$,
after which we freeze the sparsity pattern by exchanging $\tau^{[k]}_l$ with $\tau^{[K_{\texttt{p}}]}_l$, and continually fine-tune the model for another $K_{\texttt{FT}}$ iterations for performance compensation.
Note that we have $K = K_{\texttt{p}} + K_{\texttt{FT}}$, where the total training iterations $K$ equals to the sum of pruning iterations $K_{\texttt{p}}$ and the fine-tuning ones $K_{\texttt{FT}}$. There are no modifications to the backpropagation in training.

\begin{figure*}[h] 
\centering 
\includegraphics[width=\textwidth]{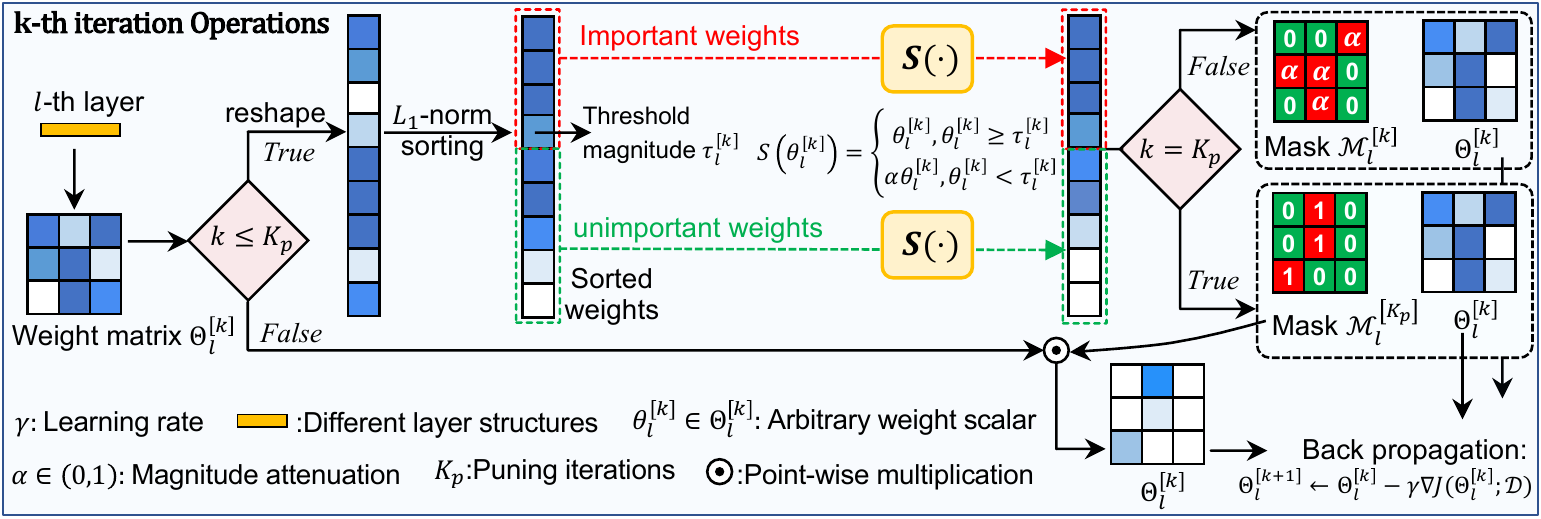}
\caption{Training pipeline of the proposed Iterative Soft Shrinkage-Percentage (ISS-P), which is exampled by $l$-th learnable layer of the network at $k$-th training iteration. 
In the pruning stage ($k\leq K_{\texttt{p}}$), ISS-P selectively attenuates the unimportant weight  and keeps the essential ones upon $L_1$-norm sorting at the forward propagation. In the fine-tuning stage ($K_{\texttt{p}}$$\leq$$ k$$\leq$$ K_{\texttt{p}}$$+$$K_{\texttt{FT}}$), the mask is frozen as $\mathcal{M}_l^{[K_{\texttt{p}}]}$ and repeatedly applied at each forward propagation. We perform a standard backpropagation in each iteration.
The proposed method can flexibly prune different types of layers (\emph{i.e.}, convolution, linear, etc.) and directly train a sparse network from the random initialization.
 }
\label{fig: framework}
\vspace{-1mm}
\end{figure*}

Different from the static mask determination, the sparse structure of IHT changes during the optimization on-the-fly, which allocates more flexibility for fitting the optimal sparse pattern. 
However, several limitations still exist. 
The first is a network throughput blocking effect. Consider the back-propagated errors between hidden layers in a neural network, $\mathbf{\delta}_{l} =  [\Theta_{l+1}^{T}\mathbf{\delta}_{l+1}]\odot \sigma'(\mathbf{z}_l)$,
where $\mathbf{\delta}_{l}$ presents the error propagated to the $l$-th layer, $\Theta_{l+1}$ denotes the weight matrix of the $(l+1)$-th layer, and $\sigma'(\mathbf{z}_l)$ computes the derivative of the activation $\sigma(\cdot)$ with the hidden representation $\mathbf{z}_l$. Due to the iterative hard thresholding operation $H(\cdot)$, there will be a certain amount of weights becomes zero, which further suspends the error transmission to the $l$-th layer, thus hindering the update of shallow weights. 
Secondly, the sparse structure of the IHT is largely susceptible to the relationship between the weight magnitude and gradient. 
The zeroed weights are vulnerable to being trapped as the ``unimportant'' category when the gradient is unexpectedly large, leading to a static sparsity during the training.
Additionally, the hard thresholding operator uniformly forces all the unimportant weight to be zeros, neglecting the inherent difference between the magnitudes.

\subsection{Iterative Soft Shrinkage}
\label{subsec: ISS}
\noindent \textbf{Iterative Soft Shrinkage-Regularization (ISS-R)}.
To address the challenges posed by the IHT, we further tailor the hard thresholding function by offering a more expressive and flexible shrinkage function toward unimportant weights, rather than solely zero-out. Accordingly, we propose a soft shrinkage function $S(\cdot)$ to facilitate the sparse network training. We first propose a regularization-driven method by introducing an $L_2$-norm regularization on weight magnitudes and implement a growing regularization~\cite{wang2020neural,bai2022dual} schedule to encourage heterogeneity among unimportant weights, namely iterative soft shrinkage-regularization (ISS-R).

Specifically, given a network with randomly initialized weights, we perform $L_1$-norm sorting to the weight magnitudes for the significant weight selection. We then include the unimportant ones into the regularization and impose an $l_2$ penalty in the backpropagation of each iteration $k\in K_{\texttt{p}}$.   Different from IHT, ISS-R naturally integrates the penalization on unimportant weights into optimization. The backpropagation of ISS-R at the pruning stage is given as
\begin{small}
\begin{equation}\label{eq: ISS-R}
    \begin{aligned}
        \theta^{[k+1]}_l\leftarrow\begin{cases} \theta^{[k]}_l - \gamma\nabla J(\theta^{[k]}_l; \mathcal{D}), \textup{~if~}\theta^{[k]}_l\geq\tau^{[k]}_l,\\ \theta^{[k]}_l\! -\! \gamma\nabla\! J(\theta^{[k]}_l; \mathcal{D})\! -\! 2\eta\theta^{[k]}_l, \textup{~if~}  \theta^{[k]}_l\!<\!\tau^{[k]}_l, \end{cases}
    \end{aligned}
\end{equation}
\end{small}where $\gamma$ is the learning rate, $\eta$ denotes the $L_2$ regularization penalization scale governed by a gradually growing schedule, \emph{i.e.}, $\eta=\eta+\delta\eta $ for every $K_\eta$ iterations, and $\delta$ controls the growing ratio of $\eta$. However, although ISS-R bypasses some limitations of the IHT, it requires tedious hyperparameter tuning (\emph{e.g.}, $\eta$, $\delta$, $K_\eta$, etc). Also, it is non-trivial to explain the effect of regularization toward the weight magnitude, leading to a sub-optimal control over the pruning. Therefore, we focus more on aligning weights with magnitude controlling and propose a novel iterative soft shrinkage by percentage in the following.

\noindent\textbf{Iterative Soft Shrinkage-Percentage (ISS-P)}.
Recall that we shrink the weight magnitude with the $L_2$ regularization in ISS-R, where the penalty intensity is proportional to the weight magnitude (see Eq.~\eqref{eq: ISS-R}). Accordingly, we can achieve a similar ISS effect by directly imposing a percentage function on weights, namely ISS-P. As shown in Fig.~\ref{fig: framework}, the training pipeline of ISS-P can be divided into two stages: 1) pruning and 2) fine-tuning.  
In the pruning stage, \emph{i.e.}, $k\leq K_{\texttt{p}}$,  the weight magnitude of the selected unimportant weights shrinks by a specific ratio. Given the $l$-th layer of the network, the soft shrinkage in forward propagation at the $k$-th iteration is formulated as
\begin{small}
\begin{equation}\label{eq: iss-p P_fwd}
    \begin{aligned}
        \theta^{[k]}_l\! = \!m_l^{[k]}\theta^{[k]}_l ~~ \textup{where}~~
        m_l^{[k]}\!=\!\begin{cases} 1, \textup{~if~}\theta^{[k]}_l\geq\tau^{[k]}_l,\\ \alpha, \textup{~if~} \theta^{[k]}_l<\tau^{[k]}_l, \end{cases}
    \end{aligned}
\end{equation}
\end{small}where we define a mask $m_l^{[k]}\in\mathcal{M}_l^{[k]}$ accounting for the weight penalization of the layer. The soft shrinkage function can be defined as $S(\cdot):=m_l^{[k]}\theta^{[k]}_l$. The $\alpha\in(0,1)$ represents the magnitude attenuation, which plays a similar role as the $\eta$ in ISS-R.  The schedule of the $\alpha$ could be customized by referring to different layers and iterations. In this work, we empirically find that setting $\alpha$ as a constant value yields a promising performance. 

\begin{algorithm}[t]
\caption{ISS-P Training} \label{algo: ISS-P}
\KwIn{train set $\mathcal{D}$, initialized parameters $\Theta$, pruning ratio $r$, total number of learnable layers $L$, \emph{i.e.}, $l \in {1,2,...,L}$. Pruning iterations $K_{\texttt{p}}$, fine tuning iterations $K_{\texttt{FT}}$, mask $\mathcal{M}_l^{[k]}=\varnothing$, magnitude attenuation $\alpha$,  learning rate $\gamma$;}
\KwOut{$\Theta$}
    \For{$k=1,...,K_{\texttt{p}}$}{
        \For{$l=1,2,...,L$}{
    Determine the $\tau_l^{[k]}$ by $L_1$-norm sorting\;
    Determine the $\mathcal{M}_l^{[k]}$\;
    Forward propagation using Eq.~\eqref{eq: iss-p P_fwd}\;
    }
    Backpropagation \begin{small}$\Theta$$\leftarrow$$\Theta$$-$$\gamma$$\nabla$$ J(\Theta;\mathcal{D})$ with Eq.~\eqref{eq: SR loss}\end{small}\;
    }
    \For{$k=K_{\texttt{p}}+1,K_{\texttt{p}}+2,...,K_{\texttt{p}}+K_{\texttt{FT}}$}{
        \For{$l=1,2,...,L$}{
        Forward propagation using Eq.~\eqref{eq: iss-p FT_fwd}\;
        }
    
    Backpropagation \begin{small}$\Theta$$\leftarrow$$\Theta$$-$$\gamma$$\nabla$$ J(\Theta;\mathcal{D})$ with Eq.~\eqref{eq: SR loss}\end{small}\;
    }
\end{algorithm}

\begin{figure}[t] 
\centering 
\includegraphics[width=.475\textwidth]{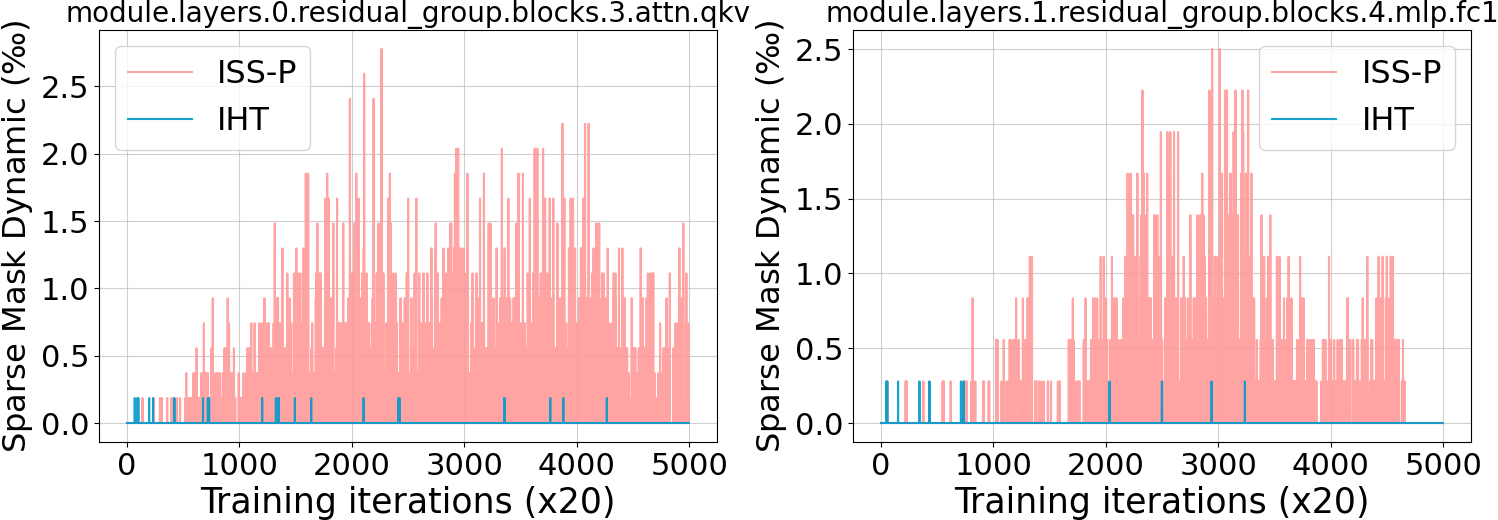}
\caption{Sparsity dynamics comparison between the ISS-P and IHT in the pruning stage. The proposed method allows a more active sparse pattern exploitation adapting to the optimization. We choose two representative layers from the SwinIR~\cite{liang2021swinir} backbone.} 
\label{fig: mask dynamics}
\vspace{-4mm} 
\end{figure}

In the fine-tuning stage $k>K_{\texttt{p}}$, we fix the sparse structure and fine-tune the network for the performance compensation, following the same procedure as IHT and ISS-R:
\begin{small}
\begin{equation}\label{eq: iss-p FT_fwd}
    \begin{aligned}
        \theta^{[k]}_l \!= \!m_l^{[K_{\texttt{p}}]}\theta^{[k]}_l ~ \textup{where}~
        m_l^{[K_{\texttt{p}}]}\!=\!\begin{cases} 1, \textup{~if~} \theta^{[k]}_l\geq\tau^{[k]}_l,\\ 0, \textup{~if~} \theta^{[k]}_l<\tau^{[k]}_l. \end{cases}
    \end{aligned}
\end{equation} 
\end{small}Per each iteration, ISS-P handles the unimportant weight adapting to its magnitude, enabling an intuitive and granular manipulation. Besides, by leveraging a percentage-based soft shrinkage function $S(\cdot)$, the sparse network evolves in a more active way, which substantially explores more sparsity possibilities throughout the optimization. Fig.~\ref{fig: mask dynamics} demonstrates this point by comparing the mask dynamics of ISS-P and IHT in the pruning stage, where we count the permille (\textperthousand) of the flips between the important/unimportant magnitudes in $\mathcal{M}_l^{[k]}$, given by two representative layers of the Transformer backbone SwinIR~\cite{liang2021swinir}, \emph{i.e.}, $13$-th layer and $44$-th layer. It can be seen that in each iteration, the number of flips counted on the IHT is quite small, and in most situations, are actually zeros. A lot more flips observed in the training process of ISS-P, \emph{e.g.}, in the $13$-th layer, flips that over $0.5$\textperthousand $ $ of the total number of the weights are observed during the optimization. 
Thereby, the sparse structure of IHT remains static in most situations yet that in ISS-P moderately changes, which allows a higher possibility to evade inferior sparse structures. 
Besides a more dynamical sparsity, the empirical evidence (see Section~\ref{subsec: analysis}) showcases that the proposed ISS-P realizes a more favorable trainability~\cite{saxe2013exact,wang2022trainability,yue2022parameter} for the sparse network, which indicates an easier convergence for the selected sparse network.
The training process of the ISS-P is summarized in Algorithm~\ref{algo: ISS-P}.

\vspace{-3mm}

\begin{table*}[t]
\begin{center}
\scalebox{0.9}{
\begin{tabular}{|c|c|cc|cc|cc|cc|cc|} 
	\hline 
	\multirow{2}{*}{Methods} & \multirow{2}{*}{Scale} & \multicolumn{2}{c|}{Set5} & 
        \multicolumn{2}{c|}{Set14} & \multicolumn{2}{c|}{B100} 
	& \multicolumn{2}{c|}{Urban100} & \multicolumn{2}{c|}{Manga109} \\
        \cline{3-4} \cline{5-6} \cline{7-8} \cline{9-10}
 	\cline{11-12} 
	 & & PSNR & SSIM & PSNR & SSIM & PSNR & SSIM & PSNR & SSIM & PSNR & SSIM \\
 \hline \hline 
         scratch  &$\times$2&  37.62& 0.9591 &33.16  &0.9141  &31.89  &0.8958  &30.83  &0.9142  & 37.69 &0.9747   \\ 
         $L_{1}$-norm~\cite{li2016pruning}&$\times$2 &37.62 &0.9591 &33.14 &0.9145 &31.90 &0.8960 &30.90 &0.9151 &37.77 &0.9749  \\
          ASSL~\cite{zhang2021aligned}&$\times$2 &37.69 &0.9593 &33.17 &0.9145 &31.93 &0.8964 &30.96 &0.9160 &37.83 &0.9751  \\
           SRP~\cite{zhang2021learning}&$\times$2 & 37.66& 0.9592& 33.20& 0.9149& 31.94& 0.8964& 31.01& 0.9165&37.88 &0.9751  \\ 
           \rowcolor{cGrey}ISS-P (ours)&$\times$2& 37.66& 0.9593& 33.22& 0.9146& 31.93& 0.8963& 31.06& 0.9169& 37.93& 0.9753 \\
           \hline \hline 
        scratch&$\times$3 &33.72 &0.9210 &29.90 &0.8342 &28.78 &0.7971 &27.06 &0.8264 &32.22& 0.9329  \\ 
         $L_{1}$-norm~\cite{li2016pruning}&$\times$3 & 33.71& 0.9209& 29.93& 0.8344&28.79 &0.7971 &27.07 &0.8266 &32.21 &0.9331  \\
          ASSL~\cite{zhang2021aligned}&$\times$3 &33.89 &0.9223 &30.00 &0.8355 &28.42 &0.7985 &27.20 &0.8305 &32.44 &0.9355  \\
           SRP~\cite{zhang2021learning}&$\times$3 & 33.86&0.9222 &29.98 &0.8353 &28.82 &0.7980 &27.19 &0.8296 &32.40&0.9347  \\ 
           \rowcolor{cGrey}ISS-P (ours)&$\times$3 &33.85 &0.9224 &30.00 &0.8358 &28.84 &0.7984 &27.26 &0.8313 &32.48 &0.9356 \\
          \hline \hline 
        scratch&$\times$4 &  31.41& 0.8821 & 28.11&0.7700&27.25&0.7255&25.16&0.7530&28.96&0.8847\\ 
         $L_{1}$-norm~\cite{li2016pruning}&$\times$4 &31.43&0.8822&28.12&0.7700&27.26&0.7256&25.16&0.7530&28.96&0.8849 \\
          ASSL~\cite{zhang2021aligned}&$\times$4 &31.50&0.8841&28.19&0.7718&27.31&0.7280&25.26&0.7583&29.20&0.8895\\
           SRP~\cite{zhang2021learning}&$\times$4 & 31.46&0.8833&28.17&0.7713&27.29&0.7269&25.25&0.7568&29.15&0.8879\\ 
          \rowcolor{cGrey}ISS-P (ours)&$\times$4  &31.60&0.8851&28.23&0.7724&27.32&0.7277&25.32&0.7593&29.28&0.8904\\
	\hline 
\end{tabular}}
\vspace{-0.2cm}
\end{center}
\caption{PSNR/SSIM comparison of the state-of-the-art methods over SwinIR under the pruning ratio of $0.9$. }
\label{tab: benchmark0.9}
\vspace{-1mm}
\end{table*}

\begin{table*}[t]
\begin{center}
\scalebox{0.9}{
\begin{tabular}{|c|c|cc|cc|cc|cc|cc|} 
	\hline 
	\multirow{2}{*}{Methods} & \multirow{2}{*}{Scale} & \multicolumn{2}{c|}{Set5} & 
        \multicolumn{2}{c|}{Set14} & \multicolumn{2}{c|}{B100} 
	& \multicolumn{2}{c|}{Urban100} & \multicolumn{2}{c|}{Manga109} \\
        \cline{3-4} \cline{5-6} \cline{7-8} \cline{9-10}
 	\cline{11-12} 
	 & & PSNR & SSIM & PSNR & SSIM & PSNR & SSIM & PSNR & SSIM & PSNR & SSIM \\
 \hline  \hline
         scratch  &$\times$2&   37.27&0.9575&32.83&0.9106&31.63&0.8923&30.04&0.9040&36.95&0.9720\\ 
         $L_{1}$-norm~\cite{li2016pruning}&$\times$2 &37.26&0.9574&32.83&0.9107&31.63&0.8924&30.07&0.9044&36.95&0.9721\\
          ASSL~\cite{zhang2021aligned}&$\times$2 &37.39&0.9581&32.92&0.9119&31.73&0.8940&30.29&0.9080&37.21&0.9732 \\
           SRP~\cite{zhang2021learning}&$\times$2 &37.41&0.9582&32.96&0.9124&31.75&0.8941&30.40 &0.9091&37.32&0.9734\\ 
           \rowcolor{cGrey}ISS-P (ours)&$\times$2& 37.46&0.9584&33.01&0.9129&31.78&0.8945&30.52&0.9105&37.43 &0.9738\\
           \hline \hline
        scratch&$\times$3 & 33.13&0.9144&29.52&0.8264&28.52&0.7899&26.40&0.8075&30.98&0.9199\\ 
         $L_{1}$-norm~\cite{li2016pruning}&$\times$3 &33.14&0.9144&29.51&0.8263&28.52&0.7899&26.39&0.8074&30.97&0.9198\\
          ASSL~\cite{zhang2021aligned}&$\times$3 &33.37&0.9172&29.64&0.8292&28.62&0.7932&26.61&0.8148&31.43&0.9254\\
           SRP~\cite{zhang2021learning}&$\times$3 &33.33&0.9167&29.65&0.8290&28.61&0.7924&26.59&0.8131&31.36&0.9241\\ 
           \rowcolor{cGrey} ISS-P (ours)&$\times$3&33.49&0.9185&29.73&0.8306&28.66&0.7939&26.75&0.8182&31.68&0.9277\\
          \hline\hline
        scratch&$\times$4 &  30.70&0.8679&27.64&0.7581&26.98&0.7154&24.56&0.7285&27.66&0.8590\\ 
         $L_{1}$-norm \cite{li2016pruning}&$\times$4 & 30.71&0.8680&27.64&0.7580&26.98&0.7154&24.57&0.7286&27.66&0.8590\\
          ASSL\cite{zhang2021aligned}&$\times$4 & 31.03&0.8748&27.83&0.7628&27.09&0.7195&24.76&0.7373&28.16&0.8701\\      
SRP\cite{zhang2021learning}&$\times$4&30.99&0.8741&27.83&0.7626&27.09&0.7193&24.79&0.7374&28.15&0.8687\\ 
          \rowcolor{cGrey}ISS-P (ours)&$\times$4  &31.16&0.8775&27.93&0.7655&27.14&0.7218&24.91&0.7436&28.44&0.8755\\
	\hline 
\end{tabular}}
\vspace{-0.2cm}
\end{center}
\caption{PSNR/SSIM comparison of the satate-of-the-art methods over SwinIR under the pruning ratio of $0.95$. }
\label{tab: benchmark0.95}
\vspace{-4mm}
\end{table*}

\begin{table*}[h]
\begin{center}
\scalebox{0.9}{
\begin{tabular}{|c|c|cc|cc|cc|cc|cc|} 
	\hline 
	\multirow{2}{*}{Methods} & \multirow{2}{*}{Scale} & \multicolumn{2}{c|}{Set5} & 
        \multicolumn{2}{c|}{Set14} & \multicolumn{2}{c|}{B100} 
	& \multicolumn{2}{c|}{Urban100} & \multicolumn{2}{c|}{Manga109} \\
        \cline{3-4} \cline{5-6} \cline{7-8} \cline{9-10}
 	\cline{11-12} 
	 & & PSNR & SSIM & PSNR & SSIM & PSNR & SSIM & PSNR & SSIM & PSNR & SSIM \\
 \hline \hline
         scratch  &$\times$2&35.34&0.9461&31.61&0.8988&30.65&0.8788&28.06&0.8722&32.22&0.9536\\ 
         $L_{1}$-norm~\cite{li2016pruning}&$\times$2 & 35.33&0.9460&31.61&0.8988&30.65&0.8789&28.06&0.8722&33.21&0.9536 \\
          ASSL~\cite{zhang2021aligned}&$\times$2 & 35.65&0.9486&31.82&0.9007&30.80&0.8808&28.30&0.8760&33.83&0.9571\\
           SRP~\cite{zhang2021learning}&$\times$2 & 35.47&0.9468&31.65&0.8992&30.70&0.8795&28.15&0.8733&33.56&0.9551\\ 
           \rowcolor{cGrey}ISS-P (ours)&$\times$2& 36.36&0.9526&32.19&0.9047&31.13&0.8853&28.89&0.8864&35.13&0.9640\\
           \hline\hline
        scratch&$\times$3 &  31.21&0.8822&28.27&0.8001&27.70&0.7670&25.01&0.7600&27.85&0.8690\\ 
         $L_{1}$-norm~\cite{li2016pruning}&$\times$3 &  31.21&0.8821&28.27&0.8001&27.70&0.7669&25.01&0.7600&27.85&0.8690\\
          ASSL~\cite{zhang2021aligned}&$\times$3 & 31.73&0.8928&28.62&0.8082&27.90&0.7733&25.29&0.7710&28.54&0.8847\\
           SRP~\cite{zhang2021learning}&$\times$3 & 31.02&0.8779&28.17&0.7968&27.65&0.7644&24.94&0.7567&27.67&0.8630\\ 
           \rowcolor{cGrey}ISS-P (ours)&$\times$3 & 32.07&0.8990&28.86&0.8133&28.06&0.7774&25.54&0.7793&29.05&0.8932\\
          \hline\hline
        scratch&$\times$4 & 29.00&0.8197&26.48&0.7219&26.29&0.6882&23.51&0.6769&25.43&0.7924\\ 
         $L_{1}$-norm \cite{li2016pruning}&$\times$4 &   29.00&0.8198&26.48&0.7219&26.29&0.6882&23.51&0.6769&25.42&0.7924\\
          ASSL~\cite{zhang2021aligned}&$\times$4 & 29.15&0.8257&26.58&0.7266&26.35&0.6917&23.58&0.6812&25.58&0.7998\\
           SRP~\cite{zhang2021learning}&$\times$4 & 28.78&0.8120&26.33&0.7147&26.21&0.6834&23.43&0.6713&25.18&0.7808\\ 
          \rowcolor{cGrey}ISS-P (ours)&$\times$4  &  29.67&0.8419&26.94&0.7373&26.55&0.6988&23.87&0.6951&26.21&0.8205\\
\hline
\end{tabular}}
\end{center}
\vspace{-2mm}
\caption{PSNR/SSIM comparison of the state-of-the-art methods over SwinIR under the pruning ratio of $0.99$. }
\label{tab: benchmark0.99}
\vspace{-2mm}
\end{table*}

\begin{figure*}[t]
\scriptsize
\centering
\resizebox{\textwidth}{!}{
\begin{tabular}{cccc}
\hspace{-2mm}
\begin{adjustbox}{valign=t}
    \begin{tabular}{c}
    \includegraphics[width=0.191\textwidth,]{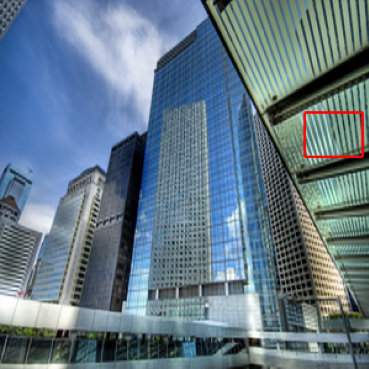}\\ 
    Urban100: img\_061 ($\times$4) \\
    \end{tabular}
\end{adjustbox}
\hspace{-2mm} 
\begin{adjustbox}{valign=t}
\begin{tabular}{ccc}
\includegraphics[width=0.17\textwidth,]{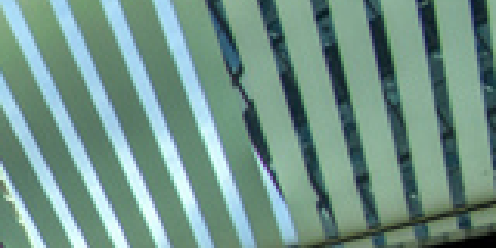}\hspace{-3mm} & 
\includegraphics[width=0.17\textwidth,]{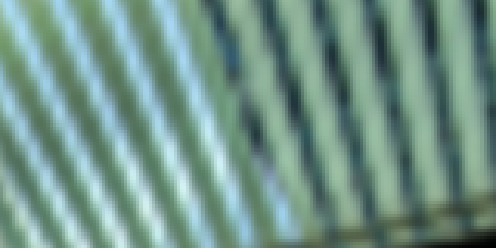}\hspace{-3mm} & 
\includegraphics[width=0.17\textwidth,]{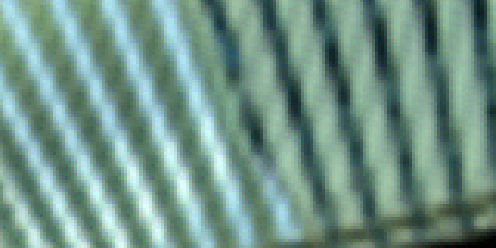}\hspace{1mm} \\
HR\hspace{-3mm} & Bicubic \hspace{-3mm} & $L_1$-norm~\cite{li2016pruning} \hspace{-3mm} \\
\includegraphics[width=0.17\textwidth,]{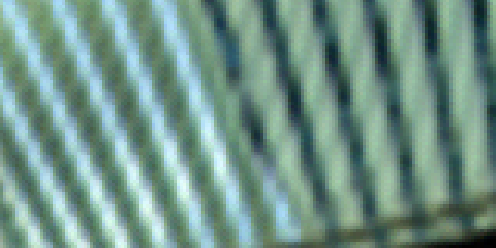}\hspace{-3mm} & 
\includegraphics[width=0.17\textwidth,]{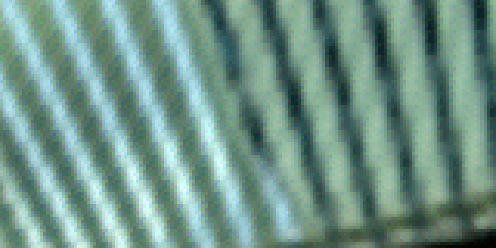}\hspace{-3mm} &  \includegraphics[width=0.17\textwidth,]{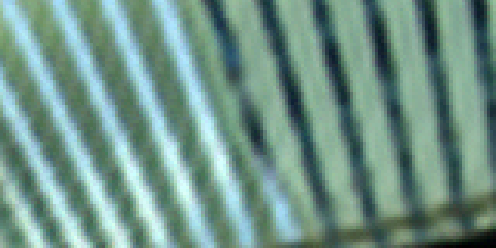}\hspace{2mm} \\
ASSL~\cite{zhang2021aligned}\hspace{-3mm} & SRP~\cite{zhang2021learning}\hspace{-3mm} & ISS-P (ours)\hspace{-3mm}\\
\end{tabular}
\end{adjustbox}
\\
\begin{tabular}{cccc}
\hspace{-2mm}
\begin{adjustbox}{valign=t}
    \begin{tabular}{c}
    \includegraphics[width=0.191\textwidth,]{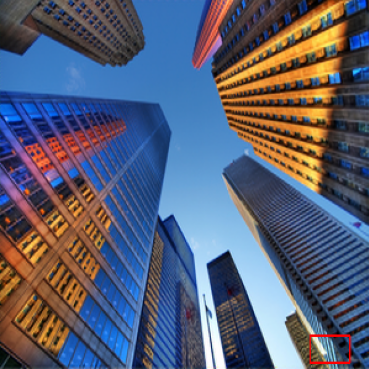}\\ 
    Urban100: img\_012 ($\times$4) \\
    \end{tabular}
\end{adjustbox}
\hspace{-2mm} 
\begin{adjustbox}{valign=t}
\begin{tabular}{ccc}
\includegraphics[width=0.17\textwidth,]{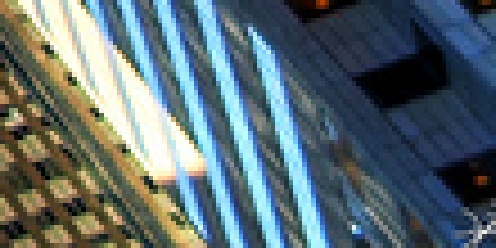}\hspace{-3mm} & 
\includegraphics[width=0.17\textwidth,]{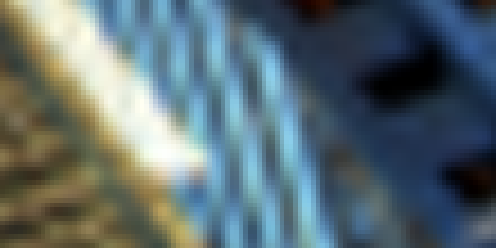}\hspace{-3mm} & 
\includegraphics[width=0.17\textwidth,]{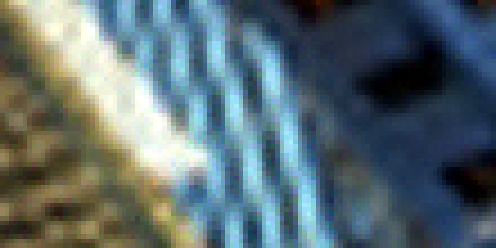}\hspace{1mm} \\
HR\hspace{-3mm} & Bicubic \hspace{-3mm} & $L_1$-norm~\cite{li2016pruning} \hspace{-3mm} \\
\includegraphics[width=0.17\textwidth,]{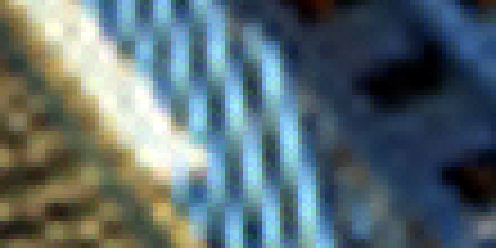}\hspace{-3mm} & 
\includegraphics[width=0.17\textwidth,]{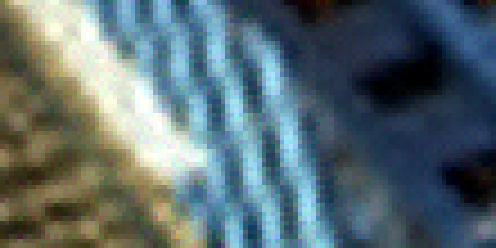}\hspace{-3mm} &  \includegraphics[width=0.17\textwidth,]{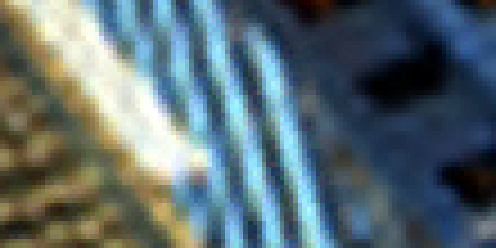}\hspace{2mm} \\
ASSL~\cite{zhang2021aligned}\hspace{-3mm} & SRP~\cite{zhang2021learning}\hspace{-3mm} & ISS-P (ours)\hspace{-3mm}\\
\end{tabular}
\end{adjustbox}
\end{tabular}

\end{tabular}}
\caption{
Visualization comparison of different pruning methods on Urban100~\cite{huang2015single} dataset. The pruning ratio is $0.99$.
}
\label{fig: visual compare 1}
\vspace{-3mm}
\end{figure*}

\begin{table*}[h]
\begin{center}
\scalebox{0.94}{
\begin{tabular}{|c|c|cc|cc|cc|cc|cc|} 
	\hline 
	\multirow{2}{*}{Backbones} & \multirow{2}{*}{Methods}  & \multicolumn{2}{c|}{Set5} & 
        \multicolumn{2}{c|}{Set14} & \multicolumn{2}{c|}{B100} 
	& \multicolumn{2}{c|}{Urban100} & \multicolumn{2}{c|}{Manga109} \\
        \cline{3-12}  
	  && PSNR & SSIM & PSNR & SSIM & PSNR & SSIM & PSNR & SSIM & PSNR & SSIM \\
 \hline 
         \multirow{5}{*}{EDSR-L} & scratch  &  29.60   &0.8522  &26.35  & 0.7312 &25.92  &0.7003 &23.69  &0.7303 &27.28  &0.8570    \\ 
         &$L_{1}$ norm \cite{li2016pruning}& 29.61 &0.8526 & 26.36& 0.7318& 25.93& 0.7011& 23.70& 0.7313& 27.35& 0.8582  \\
         &ASSL~\cite{zhang2021aligned}  &29.85 &0.8568 &26.54 &0.7368 &26.07 &0.7064 & 24.09&0.7461 &27.93 &0.8690  \\
         &SRP~\cite{zhang2021learning}  &29.78 &0.8558 &26.47 &0.7349 &26.00 &0.7036 &23.89 &0.7392 &27.72 &0.8656  \\
           &\cellcolor{cGrey}ISS-P (ours) &\cellcolor{cGrey}30.23 &\cellcolor{cGrey}0.8628 &\cellcolor{cGrey}26.74 &\cellcolor{cGrey}0.7428 &\cellcolor{cGrey}26.21 &\cellcolor{cGrey}0.7109 &\cellcolor{cGrey}24.43 &\cellcolor{cGrey}0.7596 &\cellcolor{cGrey}28.51 &\cellcolor{cGrey}0.8783  \\
        \hline
        \multirow{5}{*}{CAT-R} &scratch  &   32.19  &0.8940  &28.61 &0.7816  & 27.58 &0.7367 &26.03  &0.7846 & 30.50 &0.8902   \\ 
         &$L_{1}$ norm \cite{li2016pruning} &32.19 &0.8940 &28.59 &0.7814 &27.58 &0.7368 & 26.01& 0.7842&30.52 &0.9083  \\
         &ASSL~\cite{zhang2021aligned} & 32.08& 0.8930&28.53 &0.7803 &27.54 &0.7356 &25.90 &0.7809 &30.35 &0.9059  \\
         &SRP~\cite{zhang2021learning} & 32.24& 0.8950& 28.61& 0.7827& 27.60& 0.7382& 26.09& 0.7871&  30.61 &0.9096\\
           &\cellcolor{cGrey}ISS-P (ours) &\cellcolor{cGrey}32.66 &\cellcolor{cGrey}0.9008&\cellcolor{cGrey}28.93&\cellcolor{cGrey}0.7900&\cellcolor{cGrey}27.80&\cellcolor{cGrey}0.7444&\cellcolor{cGrey}26.94&\cellcolor{cGrey}0.8118&\cellcolor{cGrey}31.52&\cellcolor{cGrey}0.9197  \\
	\hline 
\end{tabular}}
\end{center}
\vspace{-2mm}
\caption{Performance comparison of different methods upon the representative CNN backbone, EDSR-L~\cite{lim2017enhanced}, and advanced transformer backbone, CAT-R~\cite{zheng2022cross}, at the scale of the $\times 4$. The pruning ratio is $0.95$. }
\label{tab: different backbones}
\end{table*}

\begin{table*}[t]
\begin{center}
\scalebox{0.9}{
\begin{tabular}{|c|c|cc|cc|cc|cc|cc|} 
	\hline 
	\multirow{2}{*}{Methods} & \multirow{2}{*}{Scale} & \multicolumn{2}{c|}{Set5} & 
        \multicolumn{2}{c|}{Set14} & \multicolumn{2}{c|}{B100} 
	& \multicolumn{2}{c|}{Urban100} & \multicolumn{2}{c|}{Manga109} \\
        \cline{3-12}
	 && PSNR & SSIM & PSNR & SSIM & PSNR & SSIM & PSNR & SSIM & PSNR & SSIM \\
 \hline 
         IHT&$\times$2  &37.48 &0.9585 &33.01 &0.9131 &31.78 &0.8947 &30.56 &0.9112 & 37.42 &0.9738  \\
         ISS-R&$\times$2  & 37.38& 0.9581&32.97 &0.9121 & 31.72&0.8939 & 30.33& 0.9082& 37.16& 0.9730 \\
           \rowcolor{cGrey}ISS-P&$\times$2 &37.51 &0.9587 &33.05 &0.9134 &31.82 &0.8952 &30.68 &0.9125 & 37.54&0.9741  \\
        \hline 
         IHT&$\times$3  &37.04 &0.9563 &32.64 &0.9091 &31.49 &0.8904 & 29.72& 0.8998&36.47 &0.9700  \\
         ISS-R&$\times$3 &37.07 &0.9566 &32.66 &0.9092 &31.50 &0.8906 &29.70 &0.8995 & 36.51& 0.9704 \\
           \rowcolor{cGrey}ISS-P&$\times$3 & 37.31& 0.9578&32.84 &0.9112 &31.66 &0.8929 &30.14 &0.9059 & 37.00& 0.9723 \\
        \hline  
         IHT&$\times$4 &  35.17& 0.9448&31.49 &0.8978 &30.57 &0.8781 & 27.95& 0.8740&32.91 &0.9519  \\
         ISS-R&$\times$4 & 35.37 &0.9462 &31.60 &0.8983 &30.66& 0.8790&28.10 &0.8730 &33.31 &0.9543  \\
           \rowcolor{cGrey}ISS-P&$\times$4 &  35.86&0.9496 & 31.89&0.9015 &30.87 &0.8819 &28.40 &0.8777 & 34.09& 0.9584 \\
	\hline 
\end{tabular}}
\vspace{-2mm}
\end{center}
\caption{Ablation study of different methods over SwinIR under the pruning ratio of $0.9$ at different scales. }
\label{tab: ablation}
\vspace{-2mm}
\end{table*}

\vspace{2mm}
\section{Experiment}\label{sec: experiment}

\noindent \textbf{Datasets and Backbones}.
Following the recent works~\cite{zhang2021aligned,zhang2021learning}, we use DIV2K~\cite{timofte2017ntire} and Flickr2K~\cite{lim2017enhanced} as the training datasets. Five benchmark datasets are employed for the quantitative comparison and visualization, including Set5~\cite{bevilacqua2012low}, Set14~\cite{zeyde2012single}, B100~\cite{martin2001database}, Manga109~\cite{matsui2017sketch}, and Urban100~\cite{huang2015single}. We adopt PSNR and SSIM~\cite{wang2004image} as evaluation metrics by referring Y channels in the YCbCr space. 

We train and evaluate the proposed method on representative backbones that cover convolutional network and transformer architectures: (1) SwinIR-Lightweight~\cite{liang2021swinir}, which takes a sub-pixel convolutional layer~\cite{shi2016real} for the upsampling and a convolutional layer for the final reconstruction. (2) EDSR-L~\cite{lim2017enhanced} that consists of $32$ residual blocks. (3) Cross-aggregation transformer~\cite{zheng2022cross} with regular rectangle window (CAT-R). We prune all of the learnable layers of the corresponding backbones from random initialization.

\noindent \textbf{Implementation Details}.
We conduct the same augmentation procedure as previous works~\cite{zhang2021aligned,zhang2021learning} by implementing random rotation of $90^{\circ}$, $180^{\circ}$, $270^{\circ}$, and flipping horizontally. For network training, we adopt the image patches of $64 \times 64$ with a batch size of 32. For computational efficiency, we set the batch size as $16$ for the ablation study. The training is performed upon an Adam~\cite{kingma2014adam} optimizer with $\beta_1$$=$$0.9$, $\beta_2$$=$$0.999$, and $\epsilon$$=$$10^{-8}$. The initial learning rate is $2\times e^{-4}$ with a half annealing upon every $2.5$$\times$$10^{5}$ iterations. We empirically determine the magnitude attenuation as $\alpha$$=$$0.95$. We set the total training iterations as $K$$=$$5$$\times$$10^5$ for benchmark comparison and $3\times10^5$ for the ablation study. The pruning stage is $K_{\texttt{p}}$$=$$1$$\times$$10^5$.  We implement the proposed method in PyTorch~\cite{paszke2019pytorch} on an NVIDIA RTX3090 GPU. 

\noindent \textbf{Compared Methods}.
We compare the proposed method with the classic baseline methods, \emph{i.e.}, training from scratch (dubbed as ``Scratch'') and $L_1$-norm pruning~\cite{li2016pruning} (denoted as ``$L_1$-norm''), as well as the most recent pruning practices~\cite{zhang2021aligned,zhang2021learning} dedicated to SR models. All the methods are elaborated under the unstructured pruning, and we have no pre-trained dense networks at the beginning. 
For the fairness of the comparison, we facilitate the same backbone structure, training iterations, neural network initialization, and pruning ratios for different methods. 
Among them, ASSL~\cite{zhang2021aligned} and SRP~\cite{zhang2021learning} are developed to remove the filters, but both are readily extendable to unstructured pruning.\footnote{More details could be found in supplementary.}
We keep the pruning constraints of both methods when operating on different backbones. For the proposed method, we use ISS-P as our final pruning treatment owing to its vigorous sparsity dynamics and promising performance. 

\subsection{Comparison with Advanced Pruning Methods}\label{subsec: benchmark}

\noindent \textbf{Performance Comparisons}. 
We conduct a thorough quantitative comparison with different pruning ratios, \emph{i.e.}, $0.9$, $0.95$, and $0.99$, under the scale of $\times2$, $\times3$, and $\times$4. As shown in Table~\ref{tab: benchmark0.9}$\sim$\ref{tab: benchmark0.99}, the proposed ISS-P presents a promising performance by improving existing methods with a considerable margin. 
Notably, the advantage of the ISS-P is amplified when the scale or pruning ratio raises. 
Thanks to the dedicated design of ISS-P, a more regularized gradient flow is preserved, leading to better trainability, especially for sparse networks with larger scale or pruning ratios. We also provide more analysis on convergence in Section~\ref{subsec: analysis}.

\noindent \textbf{Visual Comparisons}. 
We further visually compare the performance of the sparse networks trained with different pruning methods. In Fig.~\ref{fig: visual compare 1}, we present the results at a challenging scale setting (\emph{i.e.}, $\times4$) and very high pruning ratio (\emph{i.e.}, $0.99$). By comparison, the proposed ISS-P allows a more granular reconstruction, especially in textured areas with detailed visual ingredients, for example, the more clear contours of the buildings. Besides, the proposed method produces fewer distortions for regions with high gradients, \emph{e.g.}, by producing clearer and more consistent edges. These observations indicate a better modeling capacity, owing to an appropriate sparse architecture upon active sparse dynamics of ISS-P and a more promising optimization. 

\noindent \textbf{Different Backbones}.
In Table~\ref{tab: different backbones}, we present the effectiveness of the proposed pruning method on different backbones. The ISS-P works favorably well by outperforming baseline and prevailing methods for SR, which is consistent with the results on SwinIR-Lightweight. Results on these backbones demonstrate that the proposed method is network structure-independent, which potentially eases the deployment of advanced SR networks and is actually an important property in practice. 

\subsection{ISS Analysis}\label{subsec: analysis}

\noindent \textbf{Ablation Study}.
We perform ablation studies of the proposed method under the pruning ratio of $0.9$ at different scales.  We specifically compare three ablated pruning methods, \emph{i.e.}, IHT, ISS-R, and ISS-P, which explore dynamic sparse structures with different weight annealing operators at each forward propagation procedure. The SwinIR-Lightweight~\cite{liang2021swinir} is adopted as the backbone. As shown in Table~\ref{tab: ablation}, the ISS-P consistently outperforms on different testing datasets. For example, the performance gap between ISS-P and ISS-R remains over $0.2$dB under the scale of $\times4$. Note that ISS-R is inferior to the IHT at $\times2$ scale but surpasses it at the $\times4$. This suggests a strong resilience of ISS-R schedule in trainability preserving albeit a sub-optimal hyperparameter configuration in the growing regularization.

\begin{figure}[h]
\centering
\resizebox{0.475\textwidth}{!}{
\begin{tabular}{c}
\hspace{-2mm} 
\begin{adjustbox}{valign=t}
\begin{tabular}{c}
module.layers.0.residual\_group.blocks.2.mlp.fc1 \\ 
\includegraphics[width=0.475\textwidth]{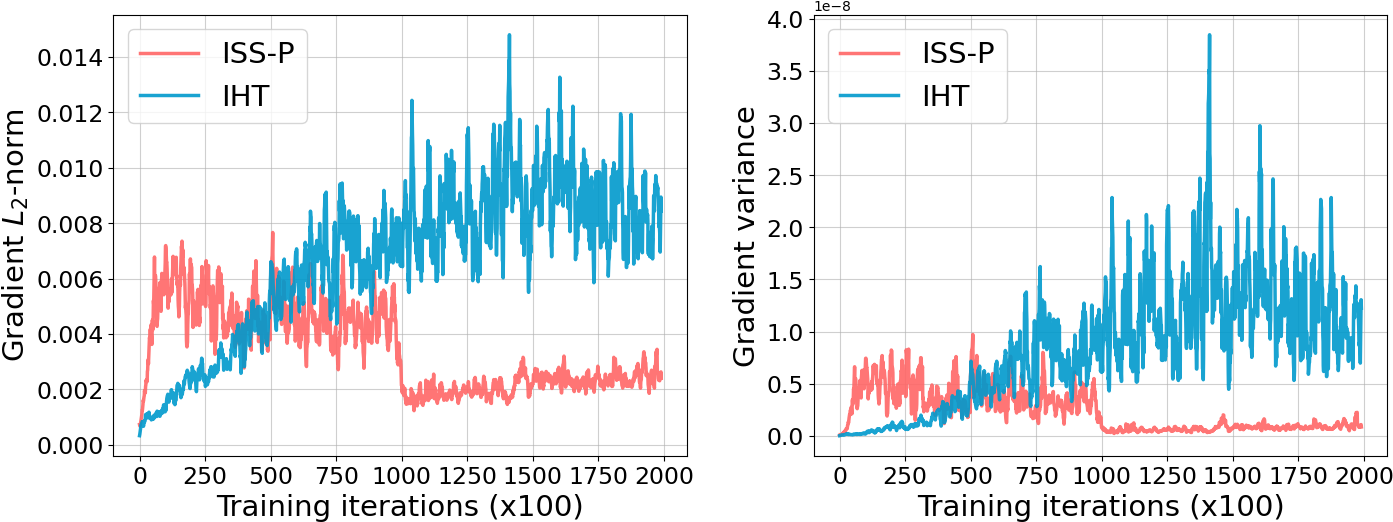} \\
module.conv\_after\_body \\ 
\includegraphics[width=0.475\textwidth]{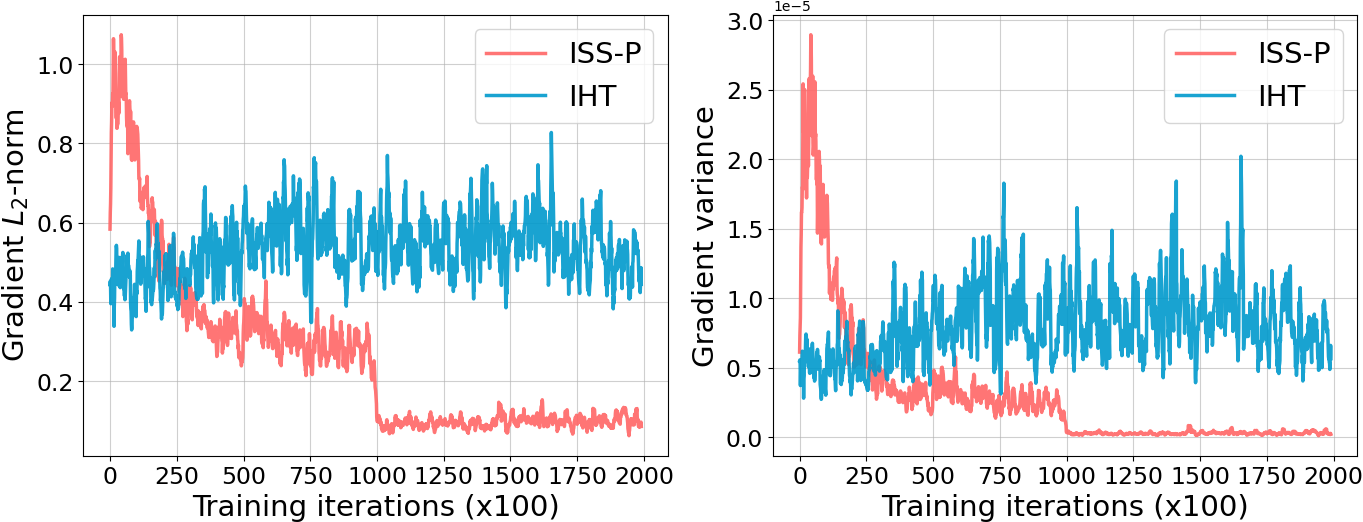}\\ 
\end{tabular}
\end{adjustbox}
\end{tabular}}
\caption{Trainability comparison of the IHT and ISS-P. The layer-wise gradient $L_2$-norm and variance in the pruning stage ($1$$\times$$10^5$ iterations) and the first $1$$\times$$10^5$ iterations of the fine-tuning stage are plotted. We choose two representative layers, \emph{i.e.}, a fully connected layer (\textit{top}) and a convolution (\textit{bottom}) from the SwinIR. 
}
\label{fig: grad stats}
\vspace{-4.5mm}
\end{figure}

\noindent \textbf{Trainability Analysis}. 
Trainability depicts whether a network is easy to be optimized, which is highly associated to the sparse structures in the field of pruning. We find that ISS-P better preserves the trainability of the network.
Intuitively, this is because the ISS-P better retains the network dynamical isometry~\cite{saxe2013exact} by better preserving the weight connections (dependencies) during the training, compared with IHT. We verify this point by observing the gradient $L_2$-norm and variance  during the training. In Fig.~\ref{fig: grad stats}, we find that the gradient norm of the ISS-P steadily converges in the pruning stage ($K_{\texttt{p}}<1\times10^5$), so that the selected sparse network (taking effect at $K_{\texttt{p}}=1\times10^5$) approaches the local minimum on the loss landscape at the end.  Reversely, the gradient descent of IHT is still ongoing (\emph{i.e.}, iteration $k=2\times10^5$), which indicates the network is harder to converge. A similar conclusion is also validated by comparing more regularized gradient variances of ISS-P against larger gradient variances of IHT. In addition, ISS-P allows better trainability throughout the network, regardless of the depth and layer types, as exampled by a shallow fully connected layer ($11$-th) and a deep convolutional layer ($101$-th).

\vspace{1.5mm}
\section{Conclusion}\label{sec: conclusion}
In this work, we have studied the problem of efficient image super-resolution by the unstructured pruning treatment upon the network with randomly initialized weights.
Specifically, we have proposed Iterative Soft Shrinkage-Percentage (ISS-P) method to
iteratively shrink the weight with a small amount proportional to the magnitude, which has 
not only enabled a more dynamic sparse structure exploitation 
but also better retained the trainability of the network.
The proposed method has been readily compatible with the off-the-shelf SR network designs, 
facilitating the sparse network acquisition and deployment.

{\small
\balance 
\bibliographystyle{ieee_fullname}
\bibliography{main}
}

\end{document}